\documentclass{article} 
\usepackage{iclr2021_conference,times}
\usepackage{listings}
\usepackage[labelfont=bf,font=small,skip=5pt]{caption}
\usepackage{hyperref}
\usepackage{url}
\usepackage{booktabs}
\usepackage{multirow}

\newcommand{\ie}{\textit{i}.\textit{e}.}
\newcommand{\eg}{\textit{e}.\textit{g}.}

\title{\centering Which Features are Learned by CodeBert: \\An Empirical Study of the BERT-based Source Code Representation Learning}

\author{Lan Zhang\thanks{equal contribution}, Chen Cao$^*$, Zhilong Wang$^*$ and Peng Liu \\
The Pennsylvania State University\\
State College, PA 16801, USA \\
\texttt{\{lfz5092,cuc96,zzw169,pxl20\}@psu.edu} \\
}

\iclrfinalcopy 
\begin{document}

\maketitle

\begin{abstract}
The Bidirectional Encoder Representations from Transformers (BERT) were proposed in the natural language process (NLP) and shows promising results. 
Recently researchers applied the BERT to source-code representation learning and reported some good news on several downstream tasks. 
However, in this paper, we illustrated that current methods cannot effectively understand the logic of source codes. The representation of source code heavily relies on the programmer-defined variable and function names.
We design and implement a set of experiments to demonstrate our conjecture and provide some insights for future works. 
\end{abstract}

\section{Introduction}
\label{sec:intro}
Deep learning has demonstrated its great learning ability in natural language processing (NLP).
To deploy a natural language task, e.g. translation and text classification, researchers first pre-train a model to embed words into vectors using ELMo ~\cite{sarzynska2021detecting}, GPT ~\cite{radford2018improving} and BERT ~\cite{devlin2018bert}. 
These pre-trained models are first learned on a large unsupervised text corpus and then fine-tuned on different downstream tasks. 
Those language-based techniques have been deployed to the source code to learn a program representation.
Similar to natural language, the program representation learned from the source code using pre-trained models can be applied for several sub-tasks for example program analysis. 
In 2020, \citeauthor{feng2020codebert} proposed a pre-trained model called CodeBERT~\cite{feng2020codebert} based on Bidirectional Encoder Representations from Transformers (BERT) that learns general-purpose representations to support downstream NL-PL applications such as natural language code search, code documentation generation, etc.
In 2021, \citeauthor{guo2020graphcodebert} proposed a new pre-trained model called GraphCodeBERT~\cite{guo2020graphcodebert}, which improves the CodeBERT by enabling the model to capture more program semantic information, such as data flow.  

The difference between natural language and program language leads to an unintended consequence if these methods are directly employed to program language.
In natural language, the meaning of a word is deterministic in a specific context, whereas in program language, a programmer can assign any string to a variable, method, or function as their name. 
In such a case, most strings in the code could be replaced by other words and may not have meaningful information.
In this case, if a BERT model still heavily relies on the literal meaning of a variable/methods/function name, it may leave a pitfall when the assigned name does not literally contain any useful information or controversial meaning.

Furthermore, limited words are used in natural language, while in the programming language, the number of words can be unlimited because a programmer can casually create a string to name a variable, no matter whether the created string is interpretable or not. Therefore, it is doubtful whether the word embedding adopted in natural language is still efficient in solving the program analysis tasks.
If a model designer ignores the numerous difference between natural language and programing language and naively adopt methods from NLP, the designed model may suffer from the above limitations.

In this paper, we aim to provide an explanation of these limitations of the BERT-based code representation learning techniques. Specifically, we want to understand what kind of features can be learned and cannot be learned by current pre-trained models. 

\lstset{
  language=C,
  frame=tb,
  basicstyle={\scriptsize \ttfamily}, 
  tabsize=3,
  breaklines=true,
  showstringspaces=false,
  numbers=left,
  numbersep=-8pt,                     
  numberstyle=\tiny\color{darkgray},
  escapeinside={(*}{*)},
  xleftmargin=2pt,
  stringstyle=\color{mauve},
  keywordstyle=\color{blue},
  commentstyle=\color{dkgreen} \textit,
  emphstyle={\color{red}},
}
\begin{lstlisting}[language=C++,label={code:goodname},caption={A piece of code with meaningful variable/function names.}, captionpos=b]{}
    template<typename It, typename Pred=std::less<typename std::iterator_traits<It>::value_type>>
    inline void bubble_sort(It begin, It end, Pred pred=Pred()){
        if ( std::distance( begin, end ) <= 1 ){ return; }
        auto it_end     = end;
        bool finished   = false;
        while ( !finished ){
            finished = true;
            std::advance( it_end, -1 );
            for (auto it = begin; it! = it_end; ++ it ){
                auto next = detail::advance( it, 1 );
                if (pred( * next, * it)){
                    std::swap( * it, * next);
                    finished = false;
                }
            }
        }   
    }
\end{lstlisting}
\begin{lstlisting}[language=C++,label={code:badname},caption={A piece of code without meaningful variable/function names.}, captionpos=b]{}
    template<typename It, typename Fun2=std::less<typename std::iterator_traits<It>::value_type>>
    inline void fun1(It var1, It var2, Pred fun2=Fun2()){
        if ( std::distance( var1, var2 ) <= 1 ){ return; }
        auto var3     = var2;
        bool var4   = false;
        while ( !var4 ){
            var4 = true;
            std::advance( var3, -1 );
            for (auto var5 = var1; var5! = var3; ++ var5 ){
                auto var6 = detail::advance( var5, 1 );
                if (fun2( * var6, * var5)){
                    std::swap( * var5, * var6);
                    var4 = false;
                }
            }
        }   
    }
\end{lstlisting}

\autoref{code:goodname} and \autoref{code:badname} are two pieces of code that achieve the same logic~\textendash~\textit{bubble sorting}.
The \autoref{code:goodname} has well-named functions and variables whereas the \autoref{code:badname} does not. 
If an analyst wants to know their purpose, through a quick glance, even a beginner can easily conclude that \autoref{code:goodname} is a bubble-sort function based on the literal meaning of the function name. 
However, it is much more challenging for an analyst to understand the purpose of \autoref{code:badname}. 
Therefore, despite the exactly the same program logic that they have, \autoref{code:badname} is much more difficult to analyze.   
We can draw the following conclusions from the analysis of these two code examples: 
1) a source code can be understood in two ways: \textit{literal analysis}, and \textit{logic analysis}.  2) The literal analysis makes a conclusion based on the name of variables and functions, which is easier to analyze but is not always reliable. 3) The logic analysis requires a high-level understanding of the code, which is more reliable but hard to analyze.   

To understand whether the existing models learn the logic of the code, 
we identify two features in the source code:
1) literal feature. 2) logic feature. For instance, a logical expression is the logic feature, whereas the variable names in the expression are literal features. 
Then, we design a set of experiments that mask out different kinds of features in the training set and observe corresponding model performance. The result shows that the current models for source code representation learning still have limited ability to learn logic features.

\section{Background}
\subsection{Deep Learning for Program Analysis}
Compared with traditional deep learning methods, researchers recognized several benefits of deep learning for the program analysis~\cite{aisecurity,CLS20}: 
First, deep learning involves less domain knowledge.  
Second, the representations learned by a DL model could be 
used for various downstream tasks.
The applications of deep learning in program analysis can be grouped into two categories:

{\em Source code level deep learning.}
CodeBert and GraphCodeBERT~\cite{feng2020codebert,guo2020graphcodebert} are pre-trained models based on Transformer which learns code representations through self-supervised training tasks ( masked language modeling and structure-aware tasks) and a large-scale unlabeled corpus. 
Specifically, CodeBERT, which is pre-trained over 6 programming languages, is trained based on three tasks: masked language modeling, code structure edge predication, and representation alignment.

{\em Assembly code level deep learning.} 
Previous research use DL to conduct various binary analysis tasks~\cite{chua2017neural,shin2015recognizing,LQY21,wang2021spotting,wang2021identifying}. 
The main focus of these works is to learn a good embedding from binary instructions or raw bytes, and then predict the label for a target task through a classification output layer. 

\section{Insights and Experiments}
A source code file of a program consists of a sequence of tokens. The tokens can be grouped into three categories: keywords, operators, and user-defined names.

Keywords are reserved words that have special meanings and purposes and can only be used for specific purposes.
For example, \texttt{for}, \texttt{if}, and \texttt{break} are widely known keywords used in many programming languages.
A programming language usually only contains a limited number of keywords. For example, \texttt{C} programming language contains 32 keywords and \texttt{Python3.7} contains 35 keywords.

Besides the keywords, a programming language needs to define a set of operators. For example, arithmetic operators (\eg, \texttt{+},  \texttt{-}, and \texttt{*}) and logical operators (\eg, \texttt{and}, \texttt{or}, and \texttt{not}) are two of most important categories.
The keywords and operators are defined by a programming language.
A programmer needs to define some tokens (\ie, names) to represent a variable, structure, function, method, class, and package. 
When programmers write a code snippet, they can randomly choose any string to name these elements. However, he/she has limited flexibility to choose the keywords and operators. Only some keywords (such as \texttt{for} and \texttt{while}), operators (such as \texttt{++}, \texttt{+1}) are exchangeable.

Currently, GraphCodeBert takes code pieces of functions or class methods as data samples. 
It tokenizes keywords, operators, and user-defined names from the code pieces. 
Inside a function or a method, we can group the user-defined names into three categories: 1) variable name. 2) method name. 3) method invocation name. 
Program logic is not affected if we map these user-defined names with other strings in the same namespace.
To evaluate whether the model learns the code semantics,
we design 4 groups of experiments. For each group of experiments, we anonymize certain categories of user-defined names. 
\begin{enumerate}
    \item In the first group of experiments, we anonymize the variable names. An example is the change from \texttt{it\_end} to \texttt{var3} and \texttt{finished} to \texttt{var4} between \autoref{code:goodname} and \autoref{code:badname}. 
    \item In the second group of experiments, we anonymize the method names. An example is the change from \texttt{bubble\_sort} to \texttt{fun1} between \autoref{code:goodname} and \autoref{code:badname}. 
    \item In the third group of experiments, we anonymize the method/function invocation names. An example is the change from \texttt{swap} to \texttt{fun2} between \autoref{code:goodname} and \autoref{code:badname}. 
    \item The last group of experiments are a combination of the first three experiments, which anonymize all three kinds of user-defined names. 
\end{enumerate}
Besides, we adopt two strategies to anonymize the name:
The first strategy called ``randomly-generated'' randomly generates strings (\eg, ``oe4yqk4cit2maq7t'') with any literal meaning. 
The second strategy called ``meaningfully-generated'' generates strings with a literal meaning. However the literal meaning does not reflect the intention of the variable/function/invocation. For example, this strategy could replace ``bubble\_sort'' with ``aes\_encryption''.


Based on the four types of name-set to replace and two replacing strategies, we eventually generated 8 variants of the original dataset from \cite{guo2020graphcodebert}. 
Then, we retrain the existing models and evaluated their performance on the existing 2 downstream tasks: natural language code search, and clone detection. 

\subsection{Experiment Results}
Figure~\ref{tab:clone} and Figure~\ref{tab:search} show experiment results (accuracy) on the downstream task of code search and code clone detection, respectively.
The second column shows the module performance reported by the original paper~\cite{guo2020graphcodebert}.
The fourth, fifth, and sixth columns show the module performance when we anonymize the variable name, method definition name, and method invocation name, respectively. The last column shows the model performance after we remove all three user-defined names.

The results show that the anonymization of the variable names, method definition names, and method invocation names will result in a huge downgrade in model performance not matter we replace user-defined names with ``randomly-generated" strings or a ``meaningfully-generated" strings.
Also, on average the dateset with meaningfully-generated strings shows worse result then the dataset with randomly-generated strings, which indicates that  ``meaningfully-generated" strings could misleading the models.
An adversarial machine learning could be trained to further exploit the weakness of the CodeBert.

Overall, our experiments proves that current source-code level representation learning methods still largely rely on the \textit{literal feature} and ignore the \textit{logic feature}. However, the \textit{literal feature} is not always reliable as mentioned in \autoref{sec:intro}. The current mode still cannot effectively learn the hidden logic feature in the source code.


\begin{table}[t]
    \centering
    \captionsetup{justification=centering}
    \footnotesize
    \caption{Results on Code Search.}
    \label{tab:search}
    \begin{tabular}{lcccccc}
    \toprule
    \bf{\emph{Language}} & \bf{\emph{Original}} & \bf{\emph{Anonymizing}} &  \bf{\emph{w/o Variable}} & \bf{\emph{w/o Method Def.}}  & \bf{\emph{w/o Method Inv.}} & \bf{\emph{All}} \\
    \midrule
    \multirow{2}{0.5cm}{Java}  & \multirow{2}{0.5cm}{70.36\%} & Random  & 67.73\% & 60.89\% & 69.84\% & 17.42\% \\
    & &  Meaningful & 67.14\%	& 58.36\% &	69.84\% &	17.03\% \\
    \midrule
    \multirow{2}{0.5cm}{Python}  & \multirow{2}{0.5cm}{68.17\%}  & Random  &  59.8\% & 55.43\% & 65.61\% & 24.09\% \\
    & & Meaningful& 59.78\% &	55.65\% &	65.61\% &	23.73\%  \\
    \bottomrule
  \end{tabular}
\end{table}

\begin{table}[t]
    \centering
    \captionsetup{justification=centering}
    \footnotesize
    \caption{Results on Clone Detection.}
    \label{tab:clone}
    \begin{tabular}{lcccccc}
    \toprule
    \bf{\emph{Language}} & \bf{\emph{Original}}  & \bf{\emph{Anonymizing}} &  \bf{\emph{w/o Variable}} & \bf{\emph{w/o Method Def.}}  & \bf{\emph{w/o Method Inv.}} & \bf{\emph{All}} \\
    \midrule
    \multirow{2}{0.5cm}{Java}  & \multirow{2}{0.5cm}{94.87\%} & Random & 92.64\% & 93.97\% & 94.72\% & 86.77\% \\
    & & Meaningful & 92.52\% &	94.27\% &	93.67\% &	84.76\% \\
    \bottomrule
  \end{tabular}
\end{table}

\subsection{Discussion}
Through a set of experiments and empirical analysis, this paper tries to explain the learning ability of current BERT-based source code representation learning schemes. The results show that CodeBERT and GraphCodeBERT are efficient to learn literal features but less efficient to learn logic features. 
 
The insights provided by this paper can help future researchers or users in two aspects:
Firstly, CodeBERT and GraphCodeBERT, which open a new area for source analysis, are efficient methods for ``well-named'' source code. However, the user and researcher should expect a lower model performance if they want to apply them to analyze source code that does not provide enough information in a variable, method, and function names, \eg, the code generated from decompilation~\cite{katz2018using} and code that does not follow standard code naming convention~\cite{butler2015investigating}.

Secondly, this paper indicates that models borrowed from NLP are not very suitable for code analysis. 
The code analysis has some significant differences compared with NLP. Logical analysis is more important in many sophisticated program analysis tasks, such as vulnerability analysis, and patching generation. But it cannot be well performed by existing model designs. It is important to investigate how to improve the model's ability for logical analysis in future research.


\bibliography{main}

\begin{thebibliography}{14}
\providecommand{\natexlab}[1]{#1}
\providecommand{\url}[1]{\texttt{#1}}
\expandafter\ifx\csname urlstyle\endcsname\relax
  \providecommand{\doi}[1]{doi: #1}\else
  \providecommand{\doi}{doi: \begingroup \urlstyle{rm}\Url}\fi

\bibitem[Butler et~al.(2015)Butler, Wermelinger, and Yu]{butler2015investigating}
Simon Butler, Michel Wermelinger, and Yijun Yu.
\newblock Investigating naming convention adherence in java references.
\newblock In \emph{2015 IEEE International Conference on Software Maintenance and Evolution (ICSME)}, pp.\  41--50. IEEE, 2015.

\bibitem[Choi et~al.(2020)Choi, Liu, Shang, Wang, Wang, Zhang, Zhou, and Zou]{CLS20}
Yoon-Ho Choi, Peng Liu, Zitong Shang, Haizhou Wang, Zhilong Wang, Lan Zhang, Junwei Zhou, and Qingtian Zou.
\newblock Using deep learning to solve computer security challenges: A survey.
\newblock \emph{Cybersecurity}, 2020.

\bibitem[Chua et~al.(2017)Chua, Shen, Saxena, and Liang]{chua2017neural}
Zheng~Leong Chua, Shiqi Shen, Prateek Saxena, and Zhenkai Liang.
\newblock {Neural Nets Can Learn Function Type Signatures from Binaries}.
\newblock In \emph{26th USENIX Security Symposium (USENIX Security 17)}, pp.\  99--116, 2017.

\bibitem[Devlin et~al.(2018)Devlin, Chang, Lee, and Toutanova]{devlin2018bert}
Jacob Devlin, Ming-Wei Chang, Kenton Lee, and Kristina Toutanova.
\newblock Bert: Pre-training of deep bidirectional transformers for language understanding.
\newblock \emph{arXiv preprint arXiv:1810.04805}, 2018.

\bibitem[Feng et~al.(2020)Feng, Guo, Tang, Duan, Feng, Gong, Shou, Qin, Liu, Jiang, et~al.]{feng2020codebert}
Zhangyin Feng, Daya Guo, Duyu Tang, Nan Duan, Xiaocheng Feng, Ming Gong, Linjun Shou, Bing Qin, Ting Liu, Daxin Jiang, et~al.
\newblock Codebert: A pre-trained model for programming and natural languages.
\newblock \emph{arXiv preprint arXiv:2002.08155}, 2020.

\bibitem[Guo et~al.(2020)Guo, Ren, Lu, Feng, Tang, Liu, Zhou, Duan, Svyatkovskiy, Fu, et~al.]{guo2020graphcodebert}
Daya Guo, Shuo Ren, Shuai Lu, Zhangyin Feng, Duyu Tang, Shujie Liu, Long Zhou, Nan Duan, Alexey Svyatkovskiy, Shengyu Fu, et~al.
\newblock Graphcodebert: Pre-training code representations with data flow.
\newblock \emph{arXiv preprint arXiv:2009.08366}, 2020.

\bibitem[Katz et~al.(2018)Katz, Ruchti, and Schulte]{katz2018using}
Deborah~S Katz, Jason Ruchti, and Eric Schulte.
\newblock Using recurrent neural networks for decompilation.
\newblock In \emph{2018 IEEE 25th International Conference on Software Analysis, Evolution and Reengineering (SANER)}, pp.\  346--356. IEEE, 2018.

\bibitem[Li et~al.(2021)Li, Qu, and Yin]{LQY21}
X.~Li, Y.~Qu, and H.~Yin.
\newblock {PalmTree: Learning an Assembly Language Model for Instruction Embedding}.
\newblock In \emph{ACM CCS}, 2021.

\bibitem[Liu et~al.(2022)Liu, Liu, Luo, Shang, Wang, Wang, Zhang, and Zou]{aisecurity}
Peng Liu, Tao Liu, Nanqing Luo, Zitong Shang, Haizhou Wang, Zhilong Wang, Lan Zhang, and Qingtian Zou.
\newblock \emph{AI for Cybersecurity: A Handbook of Use Cases}.
\newblock Amazon, 2022.
\newblock URL \url{https://www.amazon.com/gp/product/B09T3123RB/}.
\newblock Kindle edition.

\bibitem[Radford et~al.(2018)Radford, Narasimhan, Salimans, and Sutskever]{radford2018improving}
Alec Radford, Karthik Narasimhan, Tim Salimans, and Ilya Sutskever.
\newblock Improving language understanding by generative pre-training.
\newblock 2018.

\bibitem[Sarzynska-Wawer et~al.(2021)Sarzynska-Wawer, Wawer, Pawlak, Szymanowska, Stefaniak, Jarkiewicz, and Okruszek]{sarzynska2021detecting}
Justyna Sarzynska-Wawer, Aleksander Wawer, Aleksandra Pawlak, Julia Szymanowska, Izabela Stefaniak, Michal Jarkiewicz, and Lukasz Okruszek.
\newblock Detecting formal thought disorder by deep contextualized word representations.
\newblock \emph{Psychiatry Research}, 304:\penalty0 114135, 2021.

\bibitem[Shin et~al.(2015)Shin, Song, and Moazzezi]{shin2015recognizing}
Eui Chul~Richard Shin, Dawn Song, and Reza Moazzezi.
\newblock Recognizing functions in binaries with neural networks.
\newblock In \emph{24th $\{$USENIX$\}$ Security Symposium ($\{$USENIX$\}$ Security 15)}, pp.\  611--626, 2015.

\bibitem[Wang et~al.(2021{\natexlab{a}})Wang, Wang, Hu, and Liu]{wang2021identifying}
Zhilong Wang, Haizhou Wang, Hong Hu, and Peng Liu.
\newblock Identifying non-control security-critical data in program binaries with a deep neural model.
\newblock \emph{arXiv preprint arXiv:2108.12071}, 2021{\natexlab{a}}.

\bibitem[Wang et~al.(2021{\natexlab{b}})Wang, Yu, Wang, and Liu]{wang2021spotting}
Zhilong Wang, Li~Yu, Suhang Wang, and Peng Liu.
\newblock Spotting silent buffer overflows in execution trace through graph neural network assisted data flow analysis.
\newblock \emph{arXiv preprint arXiv:2102.10452}, 2021{\natexlab{b}}.

\end{thebibliography}
\bibliographystyle{iclr2021_conference}


\end{document}